%% file: revised_manuscript.tex
\documentclass[fleqn,10pt]{wlscirep}

\usepackage[utf8]{inputenc}
\usepackage[T1]{fontenc}
\usepackage{multirow}

\usepackage{xspace}
\usepackage{hyperref}

\usepackage{algorithm}
\usepackage{algorithmic}
\usepackage{amssymb}
\usepackage{amsmath}

\input{revision_commands}

\DeclareMathOperator{\sign}{sign}

\newcommand{\robustmis}{Robust-MIS 2019\xspace}
\newcommand{\ndovis}{EndoVis 2017\xspace}

\def \ie {\emph{i.e.}~}
\def \eg {\emph{e.g.}~}
\def \et {\emph{et al.}~}
\title{SegMatch: Semi-supervised surgical instrument segmentation}

\author[1]{Meng Wei}
\author[1]{Charlie Budd}
\author[1]{Luis C. Garcia-Peraza-Herrera}
\author[3]{Reuben Dorent}
\author[2]{Miaojing Shi}
\author[1]{Tom Vercauteren}
\affil[1]{School of Biomedical Engineering \& Imaging Sciences, King's College London}
\affil[2]{College of Electronic and Information Engineering, Tongji University}
\affil[3]{Harvard Medical School, Harvard University}

\begin{abstract}
Surgical instrument segmentation is recognised as a key enabler in providing advanced surgical assistance and improving computer-assisted interventions. In this work, we propose SegMatch, a semi-supervised learning method to reduce the 
need for expensive annotation for laparoscopic and robotic surgical images.
SegMatch builds on FixMatch, a widespread semi-supervised \emph{classification} pipeline combining consistency regularization and pseudo-labelling,
and adapts it for the purpose of \emph{segmentation}.
In our proposed SegMatch, the unlabelled images are first weakly augmented and fed to the segmentation model to generate pseudo-labels.
In parallel, images are fed to a strong augmentation branch and consistency between the branches is used as an unsupervised loss.
To increase the relevance of our strong augmentations, we depart from using only handcrafted augmentations and introduce a trainable adversarial augmentation strategy.
Our FixMatch adaptation for segmentation tasks further includes carefully considering the equivariance and invariance properties of the augmentation functions we rely on. 
For binary segmentation tasks,
our algorithm was evaluated on the MICCAI Instrument Segmentation Challenge datasets, \robustmis and \ndovis.
For multi-class segmentation tasks, we relied on the recent CholecInstanceSeg dataset.
\revref{3}{6}
\revmod{Our results show that SegMatch outperforms fully-supervised approaches by incorporating unlabelled data, and surpasses a range of state-of-the-art semi-supervised models across different labelled to unlabelled data ratios.}
\end{abstract}
\begin{document}

\flushbottom
\maketitle
%
%
\thispagestyle{empty}
\section*{Introduction}
\label{introduction}
\input{1_introduction.tex}

\section*{Related work}
\label{related_work}
\input{2_related_work.tex}

\section*{Method}
\label{method}
\input{3_method.tex}

\section*{Experimental setup}
\label{experiments}
\input{4_experiments.tex}

\section*{Results}
\label{results_and_discussion}
\input{5_Results_and_Discussion.tex}

\section*{Discussion and conclusion}
\label{conclusion}
\input{6_conclusion.tex}








 





\bibliography{robo}



\section*{Acknowledgements}
Meng Wei is supported by the UKRI EPSRC CDT in Smart Medical Imaging [EP/S022104/1].
This work was supported by core funding from the Wellcome Trust / EPSRC [WT203148/Z/16/Z; NS/A000049/1].
Tom Vercauteren is supported by a Medtronic / RAEng Research Chair [RCSRF1819\textbackslash7\textbackslash34].
For the purpose of open access, the authors have applied a CC BY public copyright licence to any Author Accepted Manuscript version arising from this submission.
Tom Vercauteren is a co-founder and shareholder of Hypervision Surgical.
%


\section*{Author contributions statement}
%
MW: Conceptualization, Methodology, Software, Data curation, Writing - Original draft preparation, Visualization, Investigation. 
CB: Software, Validation. 
LCGPH: Conceptualization, Writing - Review \& Editing.
RB: Software, Writing - Review \& Editing.
MS: Conceptualization, Methodology, Writing - Review \& Editing, Supervision.
TV: Conceptualization, Methodology, Writing - Review \& Editing, Supervision.

\section*{Data availability statement}

The datasets used in this study are publicly available as follows:
\begin{itemize}
    \item MICCAI Instrument Segmentation Challenge datasets:
    \begin{itemize}
    \item \robustmis - accessible at \footnotesize \url{https://www.synapse.org/Synapse:syn20575265.} \normalsize
    \item \ndovis - accessible at \footnotesize \url{https://endovissub2017-roboticinstrumentsegmentation.grand-challenge.org/}.
\end{itemize}
    \item The CholecInstanceSeg dataset:
Available at \footnotesize \url{https://www.synapse.org/Synapse:syn60239970/datasets/}.
\end{itemize}

\normalsize
These datasets are distributed under their respective open-access terms and do not require additional permissions for use.







\end{document}

%% file: revision_commands.tex
\ifdefined\showrevision

\usepackage{xcolor}
\usepackage[normalem]{ulem}
\usepackage{marginnote}
\usepackage{xspace}

\newcommand{\revref}[2]{%
\marginnote{$R_{#1}C_{#2}$}
}

\newcommand{\edtref}[1]{%
\marginnote{$EC_{#1}$}
}

\newcommand{\revmod}[1]{%
{\color{magenta}#1}\xspace
}

\newcommand{\revnew}[1]{%
{\color{orange}#1}\xspace
}

\newcommand{\revdel}[1]{%
{\color{darkgray}\sout{#1}}\xspace
}

\else

\usepackage{xspace}

\newcommand{\marginnote}[1]{\ignorespaces}
\newcommand{\revref}[2]{\ignorespaces}
\newcommand{\edtref}[1]{\ignorespaces}
\newcommand{\revmod}[1]{#1\xspace}
\newcommand{\revnew}[1]{#1\xspace}
\newcommand{\revdel}[1]{\ignorespaces}

\fi

%% file: 1_introduction.tex
\begin{figure}[!h]
\centerline{\includegraphics[width=0.5\textwidth]{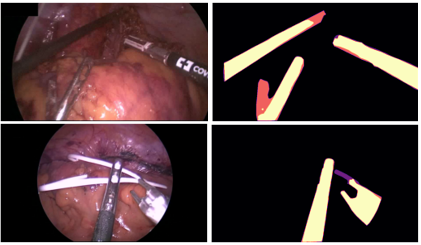}}
\caption{Representative  sample images from \robustmis of laparoscopic surgery (left) and state-of-the-art instrument segmentation results (right). True positive (yellow), true negative (black), false positive (purple), and false negative (red).}
\label{fig:open}
\end{figure}

Automatic visual understanding in
laparoscopic and robotic surgical videos is crucial for enabling autonomous surgery and providing advanced surgical support to clinical teams. Within this field,
instrument segmentation, as shown in Figure~\ref{fig:open}, is a fundamental building block. Example use cases include automatic surgical skill assessment, placing informative overlays on the display, performing augmented reality without occluding instruments, intra-operative guidance systems, surgical workflow analysis, visual servoing, and surgical task automation~\cite{ross2020robust}.
Surgical instrument segmentation has advanced rapidly from traditional methods \cite{speidel2014visual,rieke2016real}, to modern deep learning-based methods~\cite{laina2017concurrent, hasan2019u,islam2019learning,isensee2020or}. Given the significance of this task, open datasets have been released, in particular, through the organisation of computational challenges such as \robustmis~\cite{ross2020robust}.

Most studies in this area exploit
fully-supervised learning, the performance of which scales with the amount of labelled training data. 
However, given the required expertise to provide accurate manual segmentation, the collection of annotations is costly and time-consuming.
It is thus unsurprising that in comparison to industry standards for natural images, no large-scale annotated datasets for surgical tool segmentation
currently exist. This leads to significant barriers
in establishing the robustness and precision needed to deploy surgical instrument segmentation in the clinic.
To tackle this challenge, a number of weak supervision approaches
\cite{sahu2021simulation,ross2018exploiting, liu2020unsupervised, sanchez2021scalable,zhao2020learning} have been proposed which take advantage of unlabelled data or image-level labels as these are easier to capture.
While interesting conceptually, weak supervision for surgical instrument segmentation has not yet been demonstrated in practice to generalize to the largest open-access surgical datasets and achieve the required accuracy for deployment in the operating theatre.

In this work, we propose a new semi-supervised surgical instrument segmentation framework, termed as SegMatch, building upon the widely adopted semi-supervised image \emph{classification} pipeline, FixMatch~\cite{sohn2020fixmatch}.
During training, FixMatch processes unlabelled input images
through two concurrent paths which
implement weak (\eg image flip and rotation) and strong (\eg significant photometric changes) augmentations respectively.
Augmented images from both paths are then fed to a shared backbone prediction network. For regions of high confidence, the prediction from the weakly augmented image  serves as a pseudo-ground-truth against which the strongly augmented image
is compared.

In order to adapt FixMatch to the segmentation task, we make a seemingly small but critical first contribution by changing the augmentation paths in SegMatch. For classification tasks, networks are expected to be invariant with respect to all types of augmentation within a certain range tailored to the specific requirements of the task. In contrast, for segmentation tasks, networks are expected to be invariant with respect to photometric transformations (\eg contrast, brightness, hue changes) but equivariant with respect to spatial transformations (\eg rotations, translations, flips).
In SegMatch, spatial transformations that are used as augmentations are inverted after the network prediction.

Our second and main contribution to SegMatch lies in the inclusion of a learned strong augmentation strategy. Promoting prediction consistency between the weakly augmented and strongly augmented branches is what helps FixMatch and SegMatch learn from unlabelled data and generalise better.
Yet, there is no guarantee that the fixed, hand-crafted set of strong augmentation types suggested in previous work~\cite{sohn2020fixmatch} is optimal.
In fact, once the network learns to be sufficiently invariant/equivariant with respect to the fixed set of augmentation, the information learned from the unlabelled data would have saturated.
In order to guide the model to learn
continuously as the training progresses, we introduce an adversarial augmentation scheme to generate strongly augmented data. We rely on the established iterative fast gradient sign method (I-FGSM)~\cite{kurakin2018adversarial}
and integrate it into our training loop to dynamically adapt the strong augmentation as we learn from the data.

We conduct extensive experiments on the \robustmis \cite{ross2020robust} and \ndovis \cite{allan20192017} datasets for binary segmentation tasks, as well as on the CholecInstanceSeg\cite{alabi2024cholecinstancesegtoolinstancesegmentation} dataset for multi-class segmentation task.  Our study demonstrated that using a ratio of 1:9 or 3:7 of labelled data to unlabelled data within a dataset allowed our model to achieve a considerably higher mean Dice score compared to state-of-the-art semi-supervised methods. Our method shows a significant improvement in surgical instrument segmentation compared to existing fully-supervised methods by utilizing a set of 17K unlabelled images
available in the \robustmis dataset in addition to the annotated subset which most competing methods have exploited.

\revref{3}{2}
\revnew{The key contributions of this work are:
\begin{itemize}
    \item We adapt the representative semi-supervised classification pipeline with segmentation-specific transformations for weak and strong augmentations and consistency regularization.
    
    \item We propose a trainable adversarial augmentation strategy to improve the relevance of strong augmentations, ensuring the model learns robustly by leveraging both handcrafted and dynamic augmentations.

    \item Extensive experiments on the \ndovis and CholecInstanceSeg datasets demonstrate superior performance by our SegMatch compared to existing semi-supervised models for both binary and multi-class segmentation tasks.
\end{itemize}
}

%% file: 2_related_work.tex
\subsection*{Semi-supervised learning}
 
Pseudo-labelling \cite{xie2020self,qiao2018deep} is a representative approach for semi-supervised learning.
A model trained on labelled data is utilized to predict pseudo-labels for unlabelled data.
This in turn provides an extended dataset of labelled and pseudo-labelled data for training.
Consistency regularization~\cite{tarvainen2017mean,rasmus2015semi} is also a widespread technique in semi-supervised learning.
There, an auxiliary objective function is used during training to promote consistency between several model predictions, where model variability arises from techniques such as weight smoothing or noise injection.

Berthelot \et\cite{berthelot2019mixmatch} introduced MixMatch to incorporate both consistency regularization and the Entropy Minimization strategy of Grandvalet \et\cite{grandvalet2004semi} into a unified loss function for semi-supervised image classification. 
Aiming at providing a simple yet strong baseline,
Sohn \et\cite{sohn2020fixmatch} introduced FixMatch to combine consistency regularization and pseudo-labelling and achieve state-of-the-art performance on various semi-supervised learning benchmarks.

Adapting semi-supervised learning for image segmentation tasks requires dedicated strategies that account for the fact that labels are provided at the pixel level.
Previous works explored the adaptation to semantic segmentation of classical semi-supervised learning including pseudo-labelling \cite{kim2020structured,li2018semi}, and consistency regularization \cite{feng2022dmt,chen2020naive}.
Ouali \et\cite{ouali2020semi} proposed cross-consistency training (CCT) to force the consistency between segmentation predictions of unlabelled data obtained from a main decoder and those from auxiliary decoders. 
Similarly, Chen \et\cite{chen2021semi} exploited a novel consistency regularization approach called cross pseudo-supervision such that segmentation results of variously initialised models with the same input image are required to have high similarity. 
Previous work also investigated the use of generative models to broaden the set of unlabelled data from which the segmentation model can learn~\cite{souly2017semi}.
More recently, Wang et al. \cite{wang2023adversarial} proposed a contrastive learning approach for semi-supervised dense prediction, introducing adversarial negatives and an auxiliary classifier designed to track and exclude potential false negatives.

Despite such advances,
current semi-supervised semantic segmentation models derived from classification models 
have not yet demonstrated the performance gains observed with FixMatch for classification.
We hypothesise that an underpinning reason is that they
do not adequately address the issue of transformation equivariance and invariance and do not exploit modern augmentation strategies as efficiently as FixMatch.

\subsection*{Surgical instrument segmentation}
The majority of surgical instrument segmentation works are supervised methods \cite{islam2019learning,gonzalez2020isinet,qin2019surgical,jin2019incorporating,seenivasan2022global}. 
Numerous studies have explored different methods to improve the accuracy of surgical instrument segmentation. For instance, 
Islam \et\cite{islam2019learning} took advantage of task-aware saliency maps and the scan path of instruments in their multitask learning model for robotic instrument segmentation. 
By using optical flow, Jin \et\cite{jin2019incorporating} derived the temporal prior and incorporated it into an attention pyramid network to improve the segmentation accuracy. 
Gonzalez \et\cite{gonzalez2020isinet} proposed an instance-based surgical instrument segmentation network (ISINet) with a temporal consistency module that takes into account the instance’s predictions across the frames in a sequence. 
Seenivasan \et\cite{seenivasan2022global} exploited global relational reasoning for multi-task surgical scene understanding which enables instrument segmentation and tool-tissue interaction detection. 
TraSeTR \cite{zhao2022trasetr} utilizes a Track-to-Segment Transformer that intelligently exploits tracking cues to enhance surgical instrument segmentation.
Recently, large models have started to be used.
Wei et al. \cite{wei2024enhancing} for example introduced an adapter network that integrates pre-trained knowledge from foundation models into a lightweight convolutional model, enhancing both robustness and performance in surgical instrument segmentation.
Yet, the use of unlabelled task-specific data for surgical tool segmentation remains relatively untapped.

A relatively small number of
works exploited surgical instrument segmentation with limited supervision.
Jin \et\cite{jin2019incorporating} transferred predictions of unlabelled frames to that of their adjacent frames in a temporal prior propagation-based model.
Liu \et\cite{liu2020unsupervised} proposed an unsupervised approach that relies on handcrafted cues including colour, object-ness, and location to generate pseudo-labels for background tissues and surgical instruments respectively.   
Liu \et\cite{liu2021graph} introduced a graph-based unsupervised method for adapting a surgical instrument segmentation model to a new domain with only unlabelled data. 
We note that most of the existing works with limited supervision focus on exploring domain adaptation or generating different types of pseudo-labels for surgical tool segmentation models and do not exploit a FixMatch-style semi-supervised learning.

\subsection*{Adversarial learning for improved generalisation}
Deep neural networks (DNNs) have been found to be vulnerable to some well-designed input samples, known as adversarial samples. Adversarial perturbations are hard to perceive for human observers, but they can easily fool DNNs into making wrong predictions.
The study of adversarial learning concerns two aspects: 1) how to generate effective adversarial examples to attack the model~\cite{NIPS2014_5ca3e9b1}; 2) how to develop efficient defence techniques to protect the model against adversarial examples~\cite{hinton2015distilling}.  
A model which is robust to adversarial attacks is also more likely to generalise better~\cite{antoniou2017data}.
As such, we hypothesise that adversarial methods may be of relevance for semi-supervised learning.

For adversarial attacks, the earliest methods \cite{szegedy2013intriguing, NIPS2014_5ca3e9b1} rely on the gradient of the loss with respect to the input image to generate adversarial perturbations.
For instance, fast gradient sign attack (FGSM)~\cite{szegedy2013intriguing} perturbs the input along the gradient direction of the loss function to generate adversarial examples.
Tramer \et\cite{tramer2017ensemble} improved the FGSM by adding a randomization step to escape the non-smooth vicinity of the data point for better attacking.
The basic iterative method (BIM) \cite{kurakin2018adversarial}, which is also referred to as iterative FGSM (I-FGSM), improves FGSM by performing multiple smaller steps to increase the attack success rate. 
Carlin \et\cite{carlini2017towards} introduced a method now referred to as C\&W which solves a constrained optimisation problem minimising the size of the perturbation while ensuring a wrong prediction after perturbation. 
Recently, Rony \et\cite{rony2019decoupling} proposes a more efficient approach to generate gradient-based attacks with low L2 norm by decoupling the direction and norm (DDN) of the adversarial perturbation.
For adversarial defence, adversarial training \cite{madry2017towards} is a seminal method which generates adversarial examples on the fly and trains the model against them to improve the model's robustness. 
Other defence methods include relying on
generative models \cite{samangouei2018defense} and leveraging the induced randomness to mitigate the effects of adversarial perturbations in the input domain \cite{cohen2019certified}.

Although the adversarial attack has the potential to enhance the performance and robustness of deep learning models, it has not been yet applied to semi-supervised learning methods for semantic segmentation.
As we show later in this work, adversarial attacks can indeed be used to effectively augment unlabelled images used for consistency optimization.

%% file: 3_method.tex
\subsection*{Proposed SegMatch algorithm: overview}
\begin{figure*}[!ht]
\centerline{\includegraphics[width=\columnwidth]{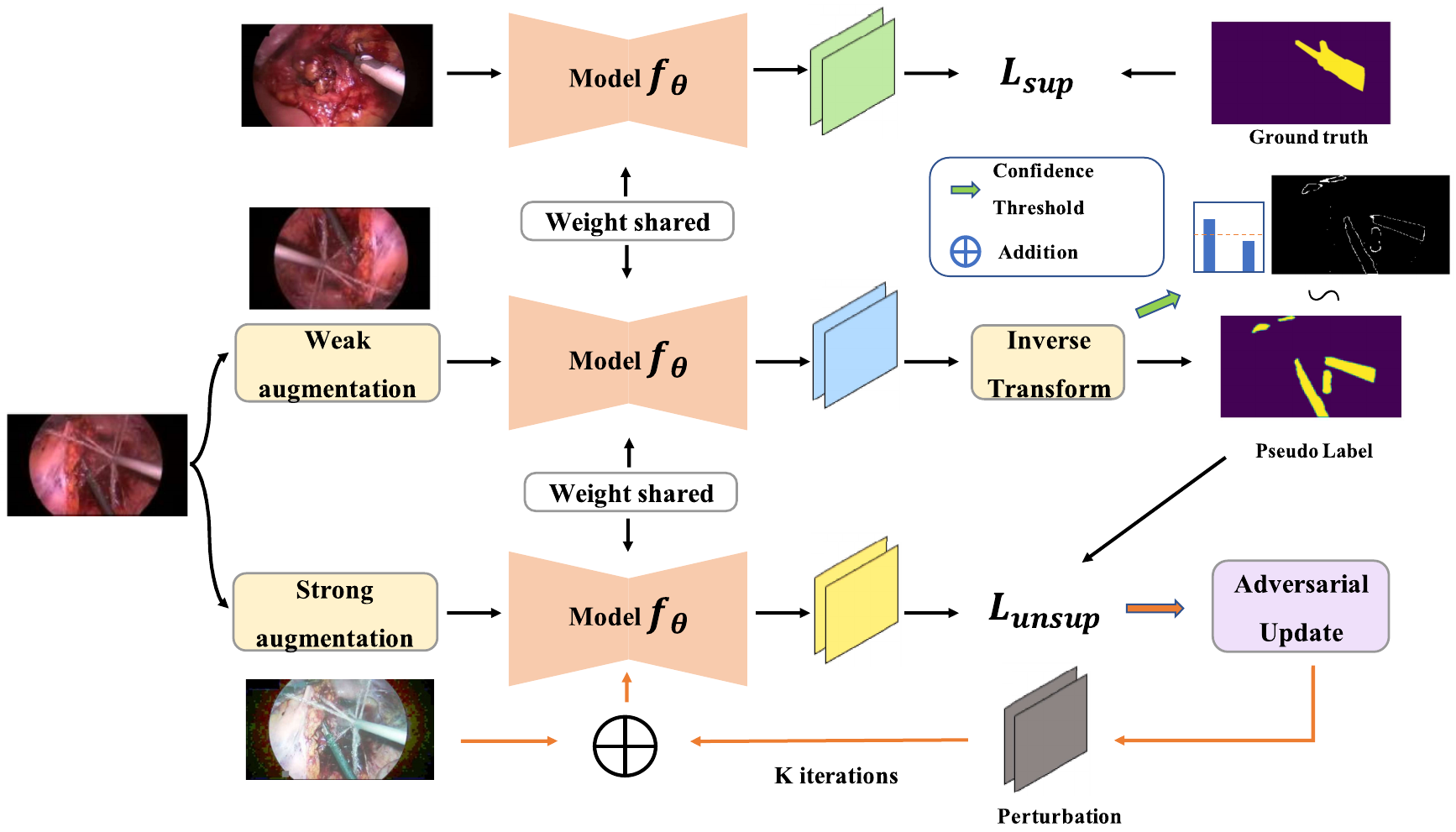}}
\caption{SegMatch training process structure.
The top row is the fully-supervised pathway which follows the traditional segmentation model training process.
The two bottom rows form the unsupervised learning pathway, where one branch uses a weakly augmented image fed into the model to compute predictions, and the second branch obtains the model prediction via strong augmentation for the same image.
The model parameters are shared across the two pathways. The hand-crafted photometric augmentation methods are used to initialize the strong augmented image, which is further perturbed by an adversarial attack (I-FGSM) for $K$ iterations.}
\label{modelstru}
\end{figure*}

Our proposed SegMatch algorithm adapts the state-of-the-art semi-supervised image \emph{classification} framework, FixMatch~\cite{sohn2020fixmatch}, to semi-supervised semantic \emph{segmentation}.
Our application primarily targets the segmentation of surgical instruments but also has the potential to be utilized for various other semantic segmentation tasks.
We follow the basic architecture of FixMatch as illustrated in Figure~\ref{modelstru}.
During training,
SegMatch uses a supervised pathway and an unsupervised pathway which share the same model parameters $\theta$ for segmentation prediction.
For the supervised pathway, the prediction from the labelled image is classically optimized against its ground truth using a standard supervised segmentation loss.
For the unsupervised pathway, given an unlabelled image, we follow FixMatch and process the input image through two concurrent paths,
which implement weak and strong augmentations respectively.
The augmented images are then fed to the model to obtain two segmentation proposals.
The output from the weak augmentation branch is designed to act as the pseudo-ground-truth for that from the strong augmentation branch.

As detailed in \hyperref[aug]{Section \emph{Weak augmentations, equivariance, pseudo-labels}}, segmentation models are typically expected to be invariant with respect to bounded photometric transformations
and equivariant with respect to spatial transformations of the input images. 
We use simple spatial transformations for our weak augmentations and propose to apply the inverse spatial transformation after the network prediction to handle spatial transformation equivariance.
We use photometric transformations for our strong augmentations and exploit a pixel-wise loss promoting consistency between the segmentation outputs of the two branches.

Given the complexity of determining the suitable range of parameters for hand-crafted strong augmentation, we propose a solution that involves a learning-based approach. As detailed in \hyperref[adv]{Section \emph{Trainable strong augmentations}},
in order to gradually increase the difficulty of the prediction from the strongly augmented branch, we introduce a learnable adversarial augmentation scheme in the strong augmentation branch.

\subsection*{Weak augmentations, equivariance, pseudo-labels}
\phantomsection
\label{aug}

\begin{figure}[bht]
\centerline{\includegraphics[width=0.6\columnwidth]{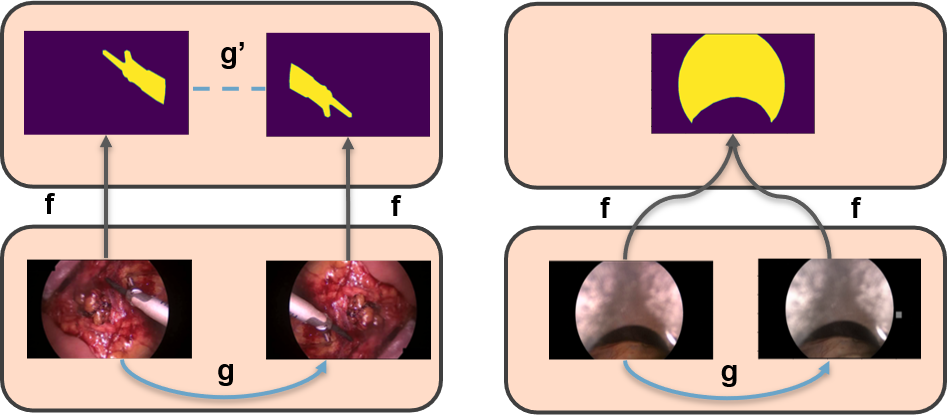}}
\caption{Equivariance (left) and invariance (right) properties for an image augmented
under different types of augmentations: spatial (left) or photometric (right).
}
\label{fig:equi}
\end{figure}
\subsubsection*{Equivariance and invariance in SegMatch}
We start by introducing the notion of equivariance and invariance
illustrated in Figure~\ref{fig:equi}.
Let us consider a function $f_\theta$ (standing for the neural network, $\theta$ being the parameters of the network) with an input $x$ (standing for the input image), and a class of transformation functions $\mathcal{G}$ (standing for a class of augmentations).
If applying $g\in\mathcal{G}$ (standing for a specific augmentation instance) to the input $x$ also reflects on the output of $f_\theta$,
that is if $f_\theta(g(x))=g(f_\theta(x))$,
then the function $f_\theta$ is said to be equivariant with respect to transformations in $\mathcal{G}$ (see Figure~\ref{fig:equi}-Left).
Conversely, if applying $g\in\mathcal{G}$ to the input $x$ does not affect the output of $f_\theta$, that is $f_\theta(g(x))=f_\theta(x)$, then the function $f_\theta$ is said to be invariant with respect to transformations in $\mathcal{G}$ (see Figure~\ref{fig:equi}-Right).

For the classification task, the model is expected to be invariant to all the augmentation strategies.
In contrast, given a segmentation model, spatial transformations on the input image should
reflect on the output segmentation map while photometric transformations should not.
Segmentation models should thus be equivariant with respect to spatial transformations and invariant for photometric transformation.
In FixMatch, weak augmentations only comprise simple
spatial transformations such as rotation, and flip, which preserve the underlying structure of the image. Meanwhile, strong augmentations only comprise photometric transformations such as posterizing and sharpness changes as provided by
the RandAugment \cite{cubuk2020randaugment} algorithm, which shifts the intensity distribution of the original image.

\subsubsection*{Inverting transformations from the weak augmentations}
Similar to FixMatch, given an input unlabelled image, we randomly select one simple spatial transformation, \ie rotation, flip, crop or resize, to apply to it in the weak augmentation branch.
In order for our SegMatch to take advantage of a consistency loss between the weak augmentation branch where spatial transformations are used (with expected equivariance of the segmentation) and the strong augmentation branch where photometric transformations are used (with expected invariance of the segmentation), we deploy an inverse spatial transformation on the output of the segmentation model in the weak augmentation branch.

Given an unlabelled image $x^u$, we denote its weak augmentation as $\omega_{e}(x^u)$. It is fed to the network $f_\theta$ to obtain the segmentation output $f_\theta(\omega_{e}(x^u))$.
We apply the inverse transformation $\omega_{e}^{-1}$ to $f_\theta(\omega_{e}(x^u))$, \ie $\omega_{e}^{-1}(f_\theta(\omega_{e}(x^u)))$.
This addresses the equivariance expectation and allows for the output of the weak augmentation branch to be easily compared
with the segmentation output from the strongly augmented image.

\subsubsection*{Soft pseudo-label generation}\label{psuedop}
Following the inverse transformation, we obtain a segmentation prediction in logit space
$p_{w}=\omega_{e}^{-1}(f_\theta(\omega_{e}(x^u)))$.
For the $i$-th pixel in $p_{w}$, \ie $p_{{w}_{i}}$, a pseudo-label is extracted by applying a sharpened softmax:
\begin{equation}
    \widetilde{y_i}=\text{Sharpen}(\text{Softmax}(p_{w_{i}}), T)
\end{equation}
where the
distribution sharpening operation of \cite{berthelot2019mixmatch} is defined as 
\begin{equation}
\text{Sharpen}(d,T)_i:=\frac{d_i^{\frac{1}{T}}}{\sum_{j=1}^{L}d_{j}^{\frac{1}{T}}}
\end{equation}
with $T$ being a temperature hyper-parameter.
The sharpening operation
allows us to control the level of confidence in the resulting probabilities.

\subsection*{Trainable strong augmentations}\label{adv}
%
%
%
%
To tackle the generalization problem
typically faced by convolutional neural networks, previous work has employed strong augmentations techniques \cite{sohn2020fixmatch, berthelot2019mixmatch, berthelot2019remixmatch}. 
However, these augmentations are hand-crafted and designing realistic strong augmentations to bridge large and complex domain gaps is challenging~\cite{garcia2021image}. 
This challenge is further exacerbated in segmentation tasks which are highly sensitive to pixel-level perturbations due to their pixel-level prediction nature~\cite{minaee2021image}.
For these reasons, despite utilizing powerful hand-crafted augmentations during training, existing methods still demonstrate limited generalization capabilities. 
In this section, we propose to tackle these key limitations by learning to perform data augmentation using adversarial perturbation during model training.  

%
%

\subsubsection*{Strong augmentation initialization}\label{stgaug}
Rather than completely replacing the strong augmentation approach in FixMatch, we build on it
to serve as initialization which will be further perturbed.
We chose the strong augmentations from the operations in RandAugment~\cite{cubuk2020randaugment} based on two criteria.
First, we focus on photometric transformations only as these satisfy the invariance expectation and do not require the use of an inverse transformation as discussed in \hyperref[aug]{Section \emph{Weak augmentations, equivariance, pseudo-labels}}.
Second, we select rather basic transformations that provide visually plausible augmentations, thereby leaving the more complex changes to the trainable refinement.
%
More specifically, our initial augmentation for the strong augmentation branch is a composition of three transformations randomly chosen from a collection of handcrafted photometric augmentation strategies.
These include adjustments to contrast, brightness, colour, and sharpness, as well as the addition of random noise, posterization, and solarization.
The strength of the individual transformations is chosen according to predefined hyper-parameters.


\subsubsection*{Adversarial augmentation approach}
As a simple yet powerful adversarial method,
we decide to use the iterative fast gradient sign method (I-FGSM) \cite{kurakin2018adversarial}, which applies multiple gradient updates by iterative small steps.


I-FGSM is based on FGSM which provides an adversarial perturbation to the input image in a single gradient-based operation.
In FGSM, the direction of the perturbation is computed from the gradient of the loss with respect to the input image.
The magnitude of the gradient is discarded by keeping only the sign along each dimension. A scaling factor is applied to keep the perturbation small. 
To compute a more refined perturbation, I-FGSM applies FGSM multiple times in smaller steps.
The perturbations are clipped
to make sure they are in the $\epsilon$-neighbourhood to the original image.
The I-FGSM equation for iteration $k+1$ out of $K$ is as follows: 
\begin{equation}
x^s_{k+1}=\text{Clip}_{x_0^s,\epsilon}\{x^s_{k}+
\frac{\epsilon}{K}
\cdot \text{Sign}(\bigtriangledown_{x^s_{k}}(L_u(f_\theta(x^s_{k}),\widetilde{y}))\}
\label{ifgsmeq}
\end{equation}
where
$x_0^s$ represents the initial strongly-augmented image;
$\widetilde{y}$ is the pseudo-label obtained from the weak augmentation branch;
$\text{Clip}\{\cdot\}$ is the clipping function which applies to every pixel in the perturbation image to limit the difference between $x^s_{{K}}$ and $x^s_{0}$ and keep it within an $\epsilon$-neighbourhood;
and
$L_u$ is the unsupervised loss function defined in equation~\eqref{eq:unsup}.
The magnitude of the perturbation $\epsilon$ and the number of I-FGSM steps $K$ are hyperparameters to adjust the quality of the adversarial approach. 


\subsection*{Loss functions in SegMatch}\label{learning}
%
In this work, the training objective for the supervised pathway is the standard pixel-wise cross-entropy loss ($l_{CE}$) combined with Dice loss ($l_{DSC}=1-DSC$), where $DSC$ represents the soft Dice coefficient:
\begin{equation}
  L_s =\frac{1}{|\mathcal D^l|}
  \sum_{x^l\in \mathcal D^l}
  \Big(
  l_{DSC}\big(y^l, f_\theta(x^l)\big)
  +
  \frac{1}{N}
  \sum_{i=0}^{N-1}
  l_{CE}\big(y^l_i, f_\theta(x^l)_i\big)
  \Big)
\end{equation}
where $x^l$ is a labelled input from the labelled set $\mathcal D^l$;
$y^l$ is the corresponding ground-truth label;
$x^l_i$ and $y^l_i$ denote the $i^{\textrm{th}}$ pixel of $x^l$ and $y^l$, respectively;
and $N$ is the number of pixels in $x^l$.

The training objective for the unsupervised pathway is a 
cross-entropy loss calculated between
a subset of confident pixel-level
pseudo-labels $\widetilde{y_i}$ stemming from the weak augmentation branch and the output probability $p_{{i}}$ from the strongly augmented image:
%
%
%
\begin{equation}
L_u = \frac{1}{{\left| D_u \right|}} \sum_{x^u\in D_u} \frac{1}{{\left| N_v^{x^u} \right|}} \sum_{i\in N_v^{x^u}} l_{CE}(\widetilde{y_i},
p_{i})
\label{eq:unsup}
\end{equation}
where
where $x^u$ is an unlabelled input from the unlabelled set $\mathcal D^u$;
$c$ denotes a specific class;
and  $N_v^{x^u}$ is the set of pixel indices with confidence score of the most confident class $\max_c(p^c_{w_{i}})$ higher than or equal to a hyper-parameter threshold $t$, i.e. $N_v^{x^u} = \{ i \, | \, \max_c(p^c_{w_{i}})\geq t\}$.

The final loss is given by:
\begin{equation}
    L=L_s+w(t)L_u
\end{equation}
where, following Laine \et\cite{laine2016temporal},
$w(t)$ is an epoch-dependent weighting function which starts from zero and ramps up along a Gaussian curve so that the supervised loss contributes to the total loss more at the beginning and the unsupervised loss increases contribution in subsequent training epochs.

\revnew{\subsection*{Data flow}}
\label{workflow}
\edtref{2}
\revnew{
To provide an overall workflow for both the weak and strong augmentation branches in SegMatch, we present detailed pseudo-code illustrations for each component in Algorithms \ref{alg:weak_augmentation} and \ref{alg:strong_augmentation}. Algorithm \ref{alg:weak_augmentation} outlines the workflow for the weak augmentation branch, while Algorithm \ref{alg:strong_augmentation} details the strong augmentation branch, which integrates hand-crafted strong augmentation initialization and adversarial augmentation.}
\revref{1}{1}
\begin{algorithm}[htbp]
\edtref{2}
\revnew{
\begin{algorithmic}
\renewcommand{\algorithmicrequire}
{\textbf{Input: Unlabelled image $X^{u}$}}
\REQUIRE
\caption{\revnew{Workflow for weak augmentation branch}}
\label{alg:weak_augmentation}
\STATE Randomly select a weak augmentation from a predefined set of spatial transformations: $\omega_e$
\STATE Apply weak augmentation to $X^{u}$: $\omega_{e}(x^u)$
\STATE Pass the augmented image through the segmentation network: $f_\theta(\omega_{e}(x^u))$
\STATE Apply the inverse spatial transformation to the output: $p_w = \omega_{e}^{-1}(f_\theta(\omega_{e}(x^u)))$
\FOR{each pixel $p_{wi}$ in $p_w$}
    \STATE Generate pseudo-label $\widetilde{y_i}$ by applying a sharpened softmax: 
    \[
    \widetilde{y_i} = \text{Sharpen}(\text{Softmax}(p_{wi}), T)
    \]
\ENDFOR
\renewcommand{\algorithmicensure}
{\textbf{Output: Model prediction of weak augmentation branch $p_w$}}
\ENSURE
\renewcommand{\algorithmicensure}
{\textbf{~~~~~~~~~~~~~~~Pseudo label $\widetilde{y}$}}
\ENSURE 
\end{algorithmic}  
}
\end{algorithm}

\begin{algorithm}[ht]
\revnew{
\begin{algorithmic}
\renewcommand{\algorithmicrequire}
{\textbf{Input: Unlabelled image $X^{u}$; Pseudo label $\widetilde{y}$; Confidence threshold $t$}}
\REQUIRE
\renewcommand{\algorithmicrequire}
{\textbf{~~~~~~~~~~~~Model prediction of weak augmentation branch $p_w$}}
\REQUIRE
\renewcommand{\algorithmicrequire}
{\textbf{~~~~~~~~~~~~Perturbation magnitude $\epsilon$; Number of steps for I-FGSM $K$}}
\REQUIRE
\renewcommand{\algorithmicensure}
{\textbf{Output: Strongly augmented image $X^{s}_{K}$}}
\caption{Workflow for strong augmentation branch}
\label{alg:strong_augmentation}
\STATE Randomly select $3$ photometric augmentation functions $\{a_1, a_2, a_3\}$ from RandAugment (set $\mathcal{A}$) and  compose the transformations: $\mathcal{T} \gets a_1 \circ a_2 \circ a_3$
\STATE Apply the composed transformation to the unlabelled image: $X_{stg} = \mathcal{T}(x^u)$
\STATE Initialization: $X_0^s=X_{stg},i = 0$
\REPEAT
\STATE Pass the augmented image through the segmentation network: $p = f_\theta(X^{s}_{i})$
\FOR{each pixel $j \in p_w$}
    \STATE Identify the most confident class: $\max_c(p^c_{w_{j}})$
    \STATE Add pixel $j$ to the confident pixel set $N_v^{x^u}$ if $max_c(p^c_{w_{j}}) \geq t$
\ENDFOR
\STATE Compute the mean loss over the confident pixels \( j \in N_v^{x^u} \):
\[
L_u^{x^u} = \frac{1}{|N_v^{x^u}|} \sum_{j \in N_v^{x^u}} l_{CE}(\widetilde{y_j}, p_j)
\]
\STATE Let $C_{\text{max}}$ the maximum allowable pixel value, e.g. 255 if unnormalised images are used
\STATE Apply one step of FGSM:\\
$X^{s}_{i+1}\leftarrow X^{s}_{i}+\frac{\epsilon}{K}\cdot\sign(\bigtriangledown x(L_u^{x^u}))$
\STATE $X_{max} \leftarrow \max\{0,X_0^s-\epsilon,X^{s}_{i+1}\}$
\STATE $X^{s}_{i+1}\leftarrow \min\{C_{max},X_0^s+\epsilon, X_{max}\}$
\STATE $i \leftarrow i + 1$
\UNTIL $i=K-1$
\ENSURE
\end{algorithmic}  
}
\end{algorithm}

%% file: 4_experiments.tex
\subsection*{Dataset}
\label{datause}
\subsubsection*{\robustmis}
\robustmis is
a laparoscopic instruments dataset including procedures in rectal resection, proctocolectomy, and sigmoid resection to detect, segment, and track medical
instruments based on endoscopic video images \cite{ross2020robust}. The training data encompasses 10-second video snippets in the form of 250 consecutive endoscopic image frames and the reference annotation for only the last frame is provided. In total, 10,040 annotated images are available from a total of 30 surgical procedures from three different types of surgery. 

As per the original challenge, the samples used for training were exclusively taken from the proctocolectomy and rectal resection procedures. These samples comprise a total of 5983 clips, with each clip having only one annotated frame while the remaining 249 frames are unannotated. 

The testing set is divided into three phases as per the original challenge, where there was no patient overlap between the training and test datasets.
\emph{Stage 1}: 325 images from the proctocolectomy procedure with another 338 images from the rectal resection procedure.
\emph{Stage 2}: 225 images from the proctocolectomy procedure and 289 others from the rectal resection procedure.
\emph{Stage 3}: 2880 annotated images from sigmoid resection, an unknown surgery which only appears in the testing stage but not in the training stage.

\subsubsection*{\ndovis}
\ndovis is
a robotic instrument image dataset captured from robotic-assisted minimally invasive surgery \cite{allan20192017}, which comprises a collection of 10 recorded sequences capturing abdominal porcine procedures. For the training phase, the first 225 frames of 8 sequences were captured at a rate of 2Hz and manually annotated with information on tool parts and types. The testing set consists of the last 75 frames from the 8 sequences used in the training data videos, along with 2 full-length sequences of 300 frames each, which have no overlap with the training phase. To prevent overlap between the training and test sets from the same surgical sequence, we followed the same split as described in ISINet \cite{gonzalez2020isinet}. This involved exclusively assigning the 2 full-length sequences for testing while keeping the training set intact with 225 frames from the remaining 8 sequences.
Note that there are no additional unannotated images for \ndovis.

\subsubsection*{CholecInstanceSeg}
The CholecInstanceSeg dataset \cite{alabi2024cholecinstancesegtoolinstancesegmentation} is a comprehensive, instance-labeled collection featuring 41,933 frames from 85 unique video sequences. This dataset includes seven distinct tool categories: Grasper, Bipolar, Hook, Clipper, Scissors, Irrigator, and Snare. Each frame has been meticulously annotated, extracted from 85 laparoscopic cholecystectomy procedures in the Cholec80 \cite{Twinanda2016EndoNetAD} and CholecT50 \cite{Nwoye2021RendezvousAM} datasets.
The dataset is divided into a training set with 55 sequences (26,830 frames), a testing set with 23 sequences (11,299 frames), and a validation set with 17 sequences (3,804 frames). In this work, we utilize the dataset solely for multi-class semantic segmentation and thus discard the instance-specific information.

\subsubsection*{Dataset usage for semi-supervised learning evaluation}
Since the above challenges were designed for fully-supervised benchmarking, we make some adaptions to evaluate SegMatch with respect to competitive semi-supervised (and fully-supervised) methods.

The \robustmis dataset was split into training and testing sets based on the original challenge splits.
The three original challenge testing stages were merged to form a single combined testing set, comprising 4057 images.
For training, we started with the full set of 5,983 labelled original challenge training images and the corresponding 17,617 unlabelled images.
%
%
In our first experiments, we use only the 5,983 labelled original challenge training images and keep only 10\%  or 30\% of the annotations. This allows for a comparison with supervised methods having access to all 5983 labelled images.
To further compare with the state-of-the-art supervised methods, we also conducted experiments using the whole 5,983 images of the training set as a labelled set, and used 17,617 additional unlabelled frames from the original videos.

For \ndovis, as no additional unlabeled data is available, the original training set, which has 1800 images in total, is split into labelled and unlabelled subsets with ratios 1:9 or 3:7.

As for the multi-class segmentation evaluation using CholecInstanceSeg, we use the same data splitting ratios as in our experiments with the \robustmis dataset.
In our ablation studies, we utilized only the original 26.8k labelled training images, retaining 30\% of the annotations.
To further evaluate multi-class segmentation performance and compare it against the baseline model, we also used the full set of 26.8k labelled training images and the 3804 frames in the validation set as the labelled dataset. Additionally, we incorporated a random selection of 66.1k unlabelled frames to assess the impact of including unlabelled data. For testing, we utilized the 11,299 frames from the original test split.

\subsection*{Implementation details}
During training, for each batch, the same number of images is sampled from the labelled dataset $D_l$ and the unlabelled dataset $D_u$. 
Within each batch, unlabelled samples are initialized by random but hand-crafted strong augmentations and then adversarially updated by adding the I-FGSM perturbations.

All our experiments were trained on two NVIDIA V100 (32GB) with a learning rate of 0.001.
The model was trained using an SGD optimizer with a momentum of 0.95, which we found can provide a smooth convergence trajectory (unreported manual tuning involved experiments with momentum values from 0.9 to 0.99). The learning rate was initialized as $1.0 \times 10^{-2}$ and decayed with the policy $lr_{ini}\times(1 - epoch/epoch_{max}) \times \eta$, where $epoch_{max}$ is the total epochs and $\eta$ is set to 0.7. The total training epoch number is 1000 and the batch size was set as 64 considering the memory constraints and training efficiency. \revref{3}{3}
\revnew{The selection of these parameters was initially based on the baseline model OR-Unet\cite{isensee2020or} and further adjusted through multiple experimental runs and prior experience with similar setups.}

\revmod{For the hyper-parameters in adversarial augmentation, the magnitude of adversarial augmentations $\epsilon$ was set to 0.08} \revref{3}{5}\revnew{based on experiments shown in Figure \ref{equi}. This value represents the optimal trade-off, which demonstrates that increasing $\epsilon$ beyond 0.08 led to performance degradation. As shown in Table \ref{tabeladvmethod} and Figure \ref{fgsmre} in the manuscript, increasing I-FGSM iterations improved segmentation performance slightly. However, considering the trade-off between performance and computational efficiency, we selected step number $K = 25$ in our final model.} Additionally, two types of initial strong augmentation techniques were utilized with pre-defined minimum and maximum magnitude values, specifically from the photometric transformations in RandAugment~\cite{cubuk2020randaugment} method.

For the segmentation model backbone, we employed the representative OR-Unet \cite{isensee2020or}.
Our OR-Unet contains 6 downsampling and 6 upsampling blocks where the residual blocks were employed in the encoder and the sequence Conv-BN-Relu layers with kernel size $3\times3$ were used in the decoder. 



\subsection*{Evaluation metrics} 
We evaluated our model based on the criterion proposed in \robustmis MICCAI challenge \cite{ross2020robust} for binary segmentation task, which includes: 
\begin{itemize}
    \item Dice Similarity Coefficient,  a widely used overlap metric in segmentation challenges;
    \item Normalized Surface Dice (NSD)~\cite{ross2020robust} is a distance-based measurement for assessing performance, which measures the overlap of two surfaces (i.e. mask borders). In adherence to the challenge guidelines \cite{ross2020robust}, we set the tolerance value for NSD to 13 pixels, which takes into account inter-rater variability. Note that the value of tolerance was determined by comparing annotations from five annotators on 100 training images as illustrated in the challenge.
\end{itemize}
For the multi-class segmentation evaluation, we adopted the evaluation method from ISINet~\cite{gonzalez2020isinet} and used three IoU-based metrics:
Ch\_IoU, ISI\_IoU, and mc\_IoU.
\begin{itemize}
    \item Ch\_IoU calculates the mean IoU for each category present in the ground truth of an image, then averages these values across all images.
    \item ISI\_IoU extends Ch\_IoU by computing the mean IoU for all predicted categories, regardless of whether they are present in the ground truth of the image. Typically, Ch\_IoU is greater than or equal to ISI\_IoU.
    \item mc\_IoU calculates the average IoU across different instrument classes and addresses category imbalance by altering the averaging order used in ISI\_IoU.
\end{itemize}

\revref{3}{6}
\revnew{
We used "percentage points" (pp) to denote the absolute change in Mean Dice score, NSD, Ch\_IoU, ISI\_IoU, and mc\_IoU, as it represents the difference between two percentage values. Additionally,} 
as shown in \hyperref[compbase]{Section \emph{Comparison with strong baselines}}, we report both the mean performance and the inter-run standard deviation (std) across folds for our proposed model, SegMatch, to capture performance consistency across multiple runs.
For other models, we provide the single-run sample-based std, which reflects consistency within a single run.
In \hyperref[abpss]{Section \emph{Ablation and parameter sensitivity study}}, we report the single-run sample-based std for both the original SegMatch model and the other ablation experiments.

%% file: 5_Results_and_Discussion.tex
\subsection*{Comparison with strong baselines}
\phantomsection
\label{compbase}
We compare our results to the state-of-the-art on \robustmis and \ndovis datasets. We categorize comparisons into two groups.
First, a head-to-head comparison is made with other semi-supervised methods.
Second, we measure the added value of incorporating unlabelled data in addition to using the complete labelled training data in \robustmis.

\subsubsection*{Comparison with semi-supervised baselines}
\begin{figure*}[!bt]
\centerline{\includegraphics[width=\linewidth]{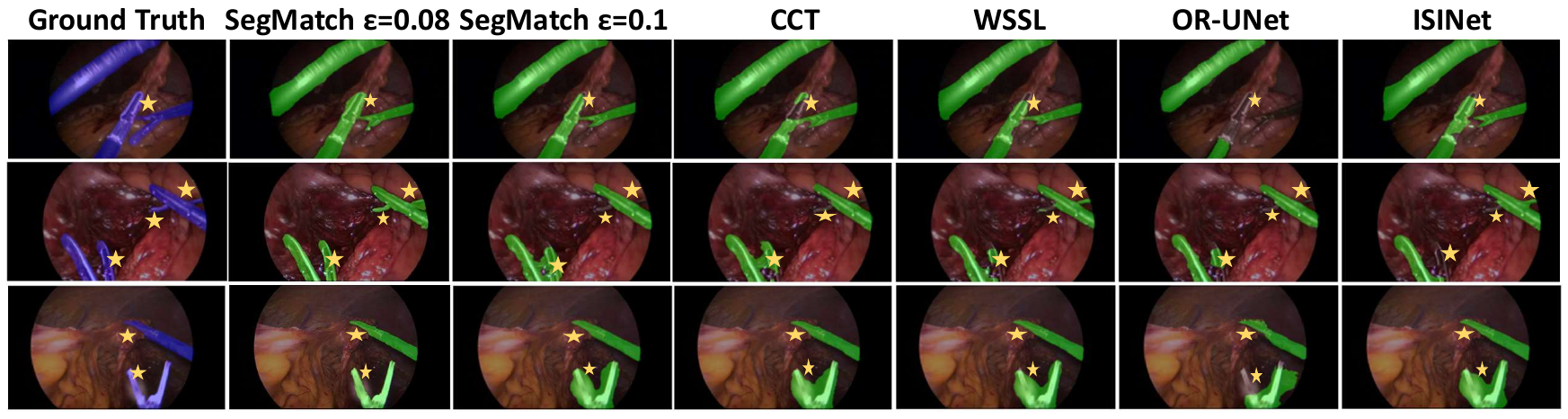}}
\revref{3}{1}
\caption{Segmentation results 
on exemplar images from three different procedures in the testing set.
Here, SegMatch, \revmod{CCT\cite{ouali2020semi}}, and \revmod{WSSL\cite{papandreou2015weakly}} were trained using the whole labelled training set of \robustmis as a labelled set, and 17K additional unlabelled frames from the original videos. The fully supervised learning models (\revmod{OR-UNet\cite{isensee2020or}} and \revmod{ISINet\cite{gonzalez2020isinet}}) were trained using the whole labelled training set of \robustmis as a labelled set.
The first column is the ground truth mask placed on top of the original image, and the other column is the segmentation results of SegMatch ablation models and state-of-the-art models. The three rows from up to button are the testing image samples from proctocolectomy procedures, sigmoid resection procedure (unseen type), and rectal resection procedure respectively. \revnew{The yellow stars highlight the key area in the  better visualization.}}
\label{segre}
\end{figure*}

For the first group, we adapted the representative semi-supervised classification method Mean-Teacher \cite{tarvainen2017mean} for the segmentation task using the same backbone network and experimental setting as ours.
We also conducted experiments with two established semi-supervised semantic segmentation models: WSSL \cite{papandreou2015weakly}, a representative benchmark in this field, and CCT \cite{ouali2020semi}, which emphasizes feature consistency across various contexts through perturbed feature alignment. Additionally, we evaluated ClassMix \cite{olsson2020classmix}, a novel data augmentation technique designed for semantic segmentation, Min-Max Similarity \cite{lou2023min}, a semi-supervised surgical instrument segmentation network based on contrastive learning, and PseudoSeg \cite{zou2021pseudoseg}, a FixMatch-based semi-supervised semantic segmentation model with a self-attention module.
Illustrative segmentation results for the selective methods on \robustmis are presented in \revref{3}{5}\revmod{Figure~\ref{segre}}, \revnew{with key areas highlighted for clarity.}

\begin{table*}[tb!]
\large
\caption{State-of-the-art semi-supervised model comparisons for \robustmis dataset (left) and \ndovis dataset (right) under differently labelled data to unlabelled data ratio. 
The reported standard deviation is measured from the performance across the testing data in a single run, except for SegMatch where we also report the standard deviation of the mean score across different runs.
\label{sttcomp}
}
\resizebox{\columnwidth}{!}{%
\begin{tabular}{l c c c c c c c c}
\toprule
\multirow{3}{*}{Methods} & \multicolumn{4}{c}{\robustmis (5983 training images)} & \multicolumn{4}{c}{\ndovis (5286 training images)}
\\\cmidrule{2-9}
&\multicolumn{2}{c}{Labelled: 598 (10\%)} & \multicolumn{2}{c}{Labelled: 1794 (30\%)} & \multicolumn{2}{c}{Labelled: 528 (10\%)} & \multicolumn{2}{c}{Labelled:1585 (30\%)}\\ 
\cmidrule{2-9}
&Mean Dice & NSD & Mean Dice & NSD & Mean Dice & NSD & Mean Dice & NSD  \\ 
\midrule
Mean-teacher \cite{tarvainen2017mean} & 62.1\(\pm\) 3.33 & 61.8 \(\pm\) 3.39 & 70.2 \(\pm\) 4.47 & 69.0 \(\pm\) 4.45 & 51.6 \(\pm\) 3.27& 52.7 \(\pm\) 3.22 & 60.2 \(\pm\) 4.28 & 59.9 \(\pm\) 4.26 \\

WSSL \cite{papandreou2015weakly} & 64.3 \(\pm\) 1.56 & 72.1 \(\pm\) 2.58 & 59.2 \(\pm\) 2.64 & 67.5 \(\pm\) 2.71 & 62.6 \(\pm\) 1.85 & 73.2 \(\pm\) 2.32 & 60.4 \(\pm\) 3.1 & 70.2 \(\pm\) 2.77
\\

CCT \cite{ouali2020semi} & 67.1 \(\pm\) 2.3 & 65.2 \(\pm\) 2.6 & 75.2 \(\pm\) 3.1 & 72.6 \(\pm\) 2.1 & 61.2 \(\pm\) 1.9 & 58.7 \(\pm\) 2.8 & 69.6 \(\pm\) 2.4 & 65.5 \(\pm\) 2.0
 \\

ClassMix \cite{olsson2020classmix} & 67.5 \(\pm\) 2.1 & 61.9 \(\pm\) 1.8 & 73.2 \(\pm\) 3.0 & 70.6 \(\pm\) 2.7 & 74.7 \(\pm\) 2.5 & 74.3 \(\pm\) 2.2 & 78.6 \(\pm\) 3.1 & 72.9 \(\pm\) 2.5
 \\

Min-Max Similarity \cite{lou2023min} & 68.7 \(\pm\) 3.13 & 61.2 \(\pm\) 2.47 & 78.6 \(\pm\) 3.29 & 72.1 \(\pm\) 2.94 & 79.4 \(\pm\) 3.18 & 77.9 \(\pm\) 2.61 & 87.1 \(\pm\) 3.36 & 76.4 \(\pm\) 3.05
  \\

PseudoSeg \cite{zou2021pseudoseg} & 70.4 \(\pm\) 2.78 & 67.5 \(\pm\) 3.01 & 79.9 \(\pm\) 2.92 & 73.4 \(\pm\) 2.64 & 79.8 \(\pm\) 3.11 & 78.7 \(\pm\) 2.85 & 87.8 \(\pm\) 3.23 & 75.5 \(\pm\) 2.99
\\

\midrule
SegMatch & 73.3 \(\pm\) 1.8 & 71.2 \(\pm\) 1.4 & 84.3 \(\pm\) 2.3 & 80.2 \(\pm\) 1.9 & 81.1 \(\pm\) 2.1 & 76.3 \(\pm\) 1.7 & 89.2 \(\pm\) 2.5 & 72.0 \(\pm\) 1.6\\
$\hookrightarrow$ std across folds & ~~~~~~~~~~\(\pm\) 0.42 & ~~~~~~~~~~ \(\pm\) 0.48 & ~~~~~~~~~~ \(\pm\) 0.61 & ~~~~~~~~~~ \(\pm\) 0.51 & ~~~~~~~~~~ \(\pm\) 0.26 & ~~~~~~~~~ \(\pm\) 0.42 & ~~~~~~~~~~ \(\pm\) 0.33 & ~~~~~~~~~~ \(\pm\) 0.45\\
\bottomrule
\end{tabular}
}
\end{table*}

Table~\ref{sttcomp} shows that, within the dataset of \robustmis, for the two labelled to unlabelled data ratios we tested, our SegMatch outperforms other methods 
with statistical significance (p$<$0.05).
Comparing SegMatch to the second-best method, PseudoSeg, we observed notable performance improvements.
Specifically, when using 10\% and 30\% labelled data in the \robustmis dataset, SegMatch achieved mean Dice score improvements of 2.9 percentage points (pp) and 4.4 pp, respectively.
Similar observations were made on the \ndovis dataset, where SegMatch outperformed PseudoSeg by 1.3 pp and 1.4 pp when utilizing 10\% and 30\% labelled data, respectively.

These results demonstrate the superior performance of SegMatch in both datasets and both labelled to unlabelled data ratios, highlighting the effectiveness of our proposed method in scenarios with limited labelled data.
%
Note that, the training dataset only consists of the proctocolectomy procedure and rectal resection procedure, and the sigmoid resection procedure is considered a new type of data for the trained model. 
Qualitatively, our proposed SegMatch is able to recognize the boundaries between different tools and segment more complete shapes for each individual tool, especially in those areas with high reflection. 

\revnew{\subsubsection*{Comparison with fully-supervised method}}
\edtref{3}
\revnew{
We compare OR-Unet \cite{isensee2020or} with our SegMatch across a wide range of labelled-to-unlabelled ratios of the \robustmis. We train the fully supervised OR-Unet \cite{isensee2020or} and our semi-supervised SegMatch using the same amount of labelled data while utilising the remaining data as unlabelled data only for SegMatch. 
} 

\revref{1}{2}
\revnew{The results shown in Table \ref{tab:fullsup-semisup} demonstrate that SegMatch outperforms OR-Unet consistently by effectively utilizing additional unlabelled data. As the amount of labelled data increases, the benefit of SegMatch diminishes, leading to the convergence of the performance of the two methods. 
Nonetheless, minor variations in the 100\% labelled case arose due to differences in codebase implementation and experimental randomness.}
\begin{table}[t]
\centering
\caption{\revnew{Performance comparison for OR-Unet \cite{isensee2020or} (fully supervised methods) vs SegMatch (our semi-supervised model) across different labelled data ratios on \robustmis dataset.}}
\label{tab:fullsup-semisup}
\resizebox{\columnwidth}{!}{
\revnew{
\begin{tabular}{l c c c c c c c c c c}
\hline
\multicolumn{11}{c}{Labelled Data Ratio} \\\cmidrule{2-11}
&\multicolumn{2}{c}{598 (10\%)} & \multicolumn{2}{c}{30\%} & \multicolumn{2}{c}{50\%} & \multicolumn{2}{c}{70\%} & \multicolumn{2}{c}{100\%}\\ 
\cmidrule{2-11}
&Mean Dice & NSD & Mean Dice & NSD & Mean Dice & NSD & Mean Dice & NSD & Mean Dice & NSD \\
\midrule
\textbf{OR-Unet \cite{isensee2020or}} & 62.1 \(\pm\) 2.9 & 60.4 \(\pm\) 2.62 &  72.6 \(\pm\) 1.76& 71.2 \(\pm\) 2.34 & 75.2 \(\pm\) 2.69 & 73.6 \(\pm\) 2.28 & 80.1 \(\pm\) 1.45 & 77.5 \(\pm\) 1.72 & 88.0 \(\pm\) 1.22 & 86.2 \(\pm\) 1.07 \\ 
\textbf{SegMatch}  & 73.3 \(\pm\) 1.89 & 71.2 \(\pm\) 1.4 &  76.3 \(\pm\) 2.3 & 74.2 \(\pm\)  1.9 & 80.9 \(\pm\) 2.65 & 78.6 \(\pm\) 1.74 & 82.7 \(\pm\) 1.42 & 79.2 \(\pm\) 2.15 & 87.5 \(\pm\) 1.09 & 84.9 \(\pm\) 0.98  \\ \hline
\end{tabular}}}
\end{table}

\subsubsection*{Added value of unlabelled data}
For the second group, we used the whole labelled training set of \robustmis
as the
labelled set and take advantage of the unlabelled video footage available from the training set of \robustmis to evaluate the impact of adding unlabelled data as mentioned in \hyperref[datause]{Section \emph{Dataset}}.

Table~\ref{stateart} shows the comparison between supervised approaches trained on the labelled set and SegMatch trained on the combination of the labelled and unlabelled sets.
Compared to the existing model OR-Unet \cite{isensee2020or},
the inclusion of additional unlabelled data in
our semi-supervised pipeline SegMatch led to a 5.7 pp higher Dice score.
Our model also demonstrates a noteworthy enhancement of 4.8 pp in comparison to the more recent ISInet \cite{gonzalez2020isinet}, 
which is now commonly employed for surgical instrument segmentation. Additionally, it maintains a 0.8 pp advantage compared to DINO-Adapter \cite{wei2024enhancing}, a recent approach that leverages pre-trained knowledge from foundation models via adapter networks
When evaluating the stage 3 testing data, which corresponds to a surgical procedure that was not seen during the training phase, our SegMatch model demonstrated superior performance compared to the official \robustmis challenge winner for stage 3 (haoyun team) by a margin of 3.9 pp.

It is noteworthy that the performance improvement from SegMatch over fully-supervised baselines is more substantial on stage 3 testing data (unseen procedure types) than on stage 1 and 2 data (procedure types represented in the training data).
This indicates that our model exhibits better generalizability.
This enhanced generalizability can enable the model to handle diverse surgical scenarios more effectively.


\begin{table*}[tb!]
\caption{Comparison on the \robustmis dataset between fully-supervised models and
SegMatch with additional unlabelled data (*~indicates only labelled data was used).
The reported standard deviation is measured from the performance across the testing data in a single run, expect for SegMatch where we also report the standard deviation of the mean score across different runs.
}
\large
\resizebox{\columnwidth}{!}{%
\begin{tabular}{l c c c c c c c c}
\toprule
\multirow{2}{*}{Method} &
\multicolumn{8}{c}{Labelled: 5983 training images (+17K unlabelled if semi-supervised)}\\\cmidrule{2-9}
&\multicolumn{2}{c}{Whole testing} & \multicolumn{2}{c}{Stage 1} & \multicolumn{2}{c}{Stage 2} & \multicolumn{2}{c}{Stage 3}\\ 
\cmidrule{2-9}
&Mean Dice & NSD & Mean Dice & NSD & Mean Dice & NSD & Mean Dice & NSD  \\
\midrule
OR-Unet \cite{isensee2020or}* & 88.0 \(\pm\) 1.22 & 86.2 \(\pm\) 1.07 & 90.2 \(\pm\) 1.15 & 88.5 \(\pm\) 1.3 & 87.9 \(\pm\) 1.1 & 85.6 \(\pm\) 1.05 & 85.9 \(\pm\) 1.2 & 84.5 \(\pm\) 1.18
\\

\robustmis winner \cite{ross2020robust}* & 90.1 \(\pm\) 1.15 & 88.9 \(\pm\) 1.08 & 92.0 \(\pm\) 1.25 & 92.7 \(\pm\) 1.1 & 90.2 \(\pm\) 1.18 & 88.6 \(\pm\) 1.05 & 89.0 \(\pm\) 1.11 & 86.4 \(\pm\) 1.09
\\

ISINet \cite{gonzalez2020isinet}* & 88.9 \(\pm\) 1.14 & 86.3 \(\pm\) 1.07 & 90.9 \(\pm\) 1.22 & 87.6 \(\pm\) 1.15 & 89.6 \(\pm\) 1.1 & 86.5 \(\pm\) 1.08 & 86.2 \(\pm\) 1.05 & 84.0 \(\pm\) 1.12\\

DINO-Adapter \cite{wei2024enhancing}* & 92.9 \(\pm\) 1.01 & 91.5 \(\pm\) 0.97 & 94.2 \(\pm\) 1.08 & 92.4 \(\pm\) 1.12 & 92.6 \(\pm\) 0.94 & 91.4 \(\pm\) 1.17 & 91.9 \(\pm\) 1.09 & 90.7 \(\pm\) 0.93
\\\midrule

SegMatch & 93.7 \(\pm\) 1.02 & 93.6 \(\pm\) 0.99 & 95.1 \(\pm\) 1.25 & 95.5 \(\pm\) 1.18 & 93.1 \(\pm\) 1.35 & 92.5 \(\pm\) 1.07 & 92.9 \(\pm\) 1.27 & 92.8 \(\pm\) 0.98\\

$\hookrightarrow$ std across folds& ~~~~~~~ \(\pm\) 0.28 & ~~~~~~~ \(\pm\) 0.24 & ~~~~~~~ \(\pm\) 0.23 & ~~~~~~~ \(\pm\) 0.19 & ~~~~~~~ \(\pm\) 0.49 & ~~~~~~~ \(\pm\) 0.31& ~~~~~~~\(\pm\) 0.12 & ~~~~~~~ \(\pm\) 0.22\\

\bottomrule
\end{tabular}}
\label{stateart}
\end{table*}

\subsubsection*{Multi-class segmentation}
To evaluate the multi-class segmentation capabilities of our model, we conducted experiments using the CholecInstanceSeg dataset. We utilized the entire labelled training set and leveraged the available unlabeled video footage to analyze the impact of incorporating unlabelled data, as discussed in \hyperref[datause]{Section \emph{Dataset}}.

Table \ref{multi-class-sta} compares our method with the baseline OR-Unet approach \cite{isensee2020or}. By incorporating an additional 66.1k unlabeled data points, our model achieved significant improvements, surpassing OR-Unet by 24.2 in Ch\_IoU, 21.89 in ISI\_IoU, and 25.06 in mc\_IoU. Additionally, our approach outperforms OR-Unet across multiple instrument categories, highlighting the effectiveness of utilizing unlabeled data in enhancing multi-class segmentation compared to the fully-supervised OR-Unet.

\begin{table*}[tb!]
\centering
\caption{Comparison of our method with state-of-the-art methods on the CholecInstanceSeg dataset for multi-class segmentation. (*~indicates only labelled data was used).
The reported standard deviation is measured from the performance across the testing data in a single run, expect for SegMatch where we also report the standard deviation of the mean score across different runs.
}
\resizebox{\linewidth}{!}{
\label{multi-class-sta}
\begin{tabular}{lcccccccccc}
\toprule
\multirow{2}{*}{Method} &
\multicolumn{9}{c}{Labelled: 26.8K training images (66.1K unlabelled if semi-supervised)}\\\cmidrule{2-11}
&Ch\_IoU & ISI\_IoU & Grasper & Bipolar
 & Hook & Clipper & Scissors & Irrigator & Snare & mc\_IoU  \\
\midrule
OR-Unet* \cite{isensee2020or} & 65.62 \(\pm\) 1.23 & 62.20 \(\pm\) 1.45 & 48.70 \(\pm\) 1.58 & 48.50 \(\pm\) 1.36 & 60.09 \(\pm\) 1.72 & 37.43 \(\pm\) 1.42 & 12.01 \(\pm\) 1.11 & 38.72 \(\pm\) 1.67 & 22.56 \(\pm\) 1.29 & 38.96 \(\pm\) 1.54
\\\midrule
SegMatch & 89.82 \(\pm\) 1.43 & 84.09 \(\pm\) 1.28 & 80.12 \(\pm\) 1.67 & 79.74 \(\pm\) 1.15 & 72.29 \(\pm\) 1.53 & 46.96 \(\pm\) 1.21 & 32.92 \(\pm\) 1.05 & 72.12 \(\pm\) 1.62 & 63.96 \(\pm\) 1.47 & 64.02 \(\pm\) 1.38\\
$\hookrightarrow$ std across folds& ~~~~~~~~~ \(\pm\) 0.31\ & ~~~~~~~~~  \(\pm\) 0.32 & ~~~~~~~~~  \(\pm\) 0.42 & ~~~~~~~~~  \(\pm\) 0.37 & ~~~~~~~~~  \(\pm\) 0.52& ~~~~~~~~~  \(\pm\) 0.27 & ~~~~~~~~~  \(\pm\) 0.34 & ~~~~~~~~~  \(\pm\) 0.43 & ~~~~~~~~~  \(\pm\) 0.26 & ~~~~~~~~~  \(\pm\) 0.29\\
\bottomrule
\end{tabular}
}
\end{table*}

\subsection*{Ablation and parameter sensitivity study}
\phantomsection
\label{abpss}
%
To evaluate the contribution of the different components of our pipeline,
we conducted ablation and parameter sensitivity studies on different strong augmentation strategies and adversarial strong augmentation methods.

\begin{table}
\centering
\caption{Ablation study results for different components, evaluating on \robustmis with different labelled and unlabelled data amount, on binary segmentation.}\label{abstdy}

\resizebox{0.7\columnwidth}{!}{%
\begin{tabular}{l c c c c}
\toprule
\multirow{2}{*}{Method} & \multicolumn{2}{c}{Labelled: 1794 (30\%)} & \multicolumn{2}{c}{Labelled: 5983 (+17K unlabelled)} \\ 
\cmidrule{2-5}
  & Mean Dice & NSD & Mean Dice & NSD   \\ 
\midrule
SegMatch & 84.3 \(\pm\) 2.3 & 80.2 \(\pm\) 1.9 & 93.7 \(\pm\) 1.02 & 93.6 \(\pm\) 0.99 \\


No adversarial & 80.6 \(\pm\) 1.67 & 77.9 \(\pm\) 1.74 & 90.6 \(\pm\) 1.87 & 89.9 \(\pm\) 1.96  \\

No adversarial \& No weak aug & 75.6 \(\pm\) 1.62 & 73.7 \(\pm\) 1.76 & 88.7 \(\pm\) 1.59 & 87.6 \(\pm\) 1.69 \\

No adversarial \& No strong aug & 71.5 \(\pm\) 1.45 & 68.2 \(\pm\) 1.37 & 85.5 \(\pm\) 1.39 & 86.9 \(\pm\) 1.42 \\

\bottomrule
\end{tabular}}
\end{table}

\begin{table}
\centering
\caption{Ablation study results for different components, evaluating on CholecInstanceSeg with different labelled and unlabelled data amount, on multi-class segmentation.}\label{abstdy-multi}

\resizebox{0.7\columnwidth}{!}{%
\begin{tabular}{l c c c c}
\toprule
\multirow{2}{*}{Method} & \multicolumn{2}{c}{Labelled: 8.04K (30\%)} & \multicolumn{2}{c}{Labelled: 26.8K (+66.1K unlabelled)} \\ 
\cmidrule{2-5}
  & Ch\_IoU & ISI\_IoU & Ch\_IoU & ISI\_IoU  \\ 
\midrule
SegMatch & 72.12 \(\pm\) 2.49 & 69.04 \(\pm\) 2.52& 89.92 \(\pm\) 1.43 & 84.09 \(\pm\) 1.28 \\


No adversarial & 67.34 \(\pm\) 1.97 & 61.27 \(\pm\) 1.92 & 84.25 \(\pm\) 1.86 & 81.29 \(\pm\) 1.78  \\

No adversarial \& No weak aug & 55.24 \(\pm\) 1.92 & 53.27 \(\pm\) 1.99 & 80.26 \(\pm\) 1.79 & 77.06 \(\pm\) 1.72 \\

No adversarial \& No strong aug & 54.32 \(\pm\) 1.66 & 52.66 \(\pm\) 1.64 & 76.90 \(\pm\) 1.77 & 72.34 \(\pm\) 1.75 \\

\bottomrule
\end{tabular}}
\end{table}


\subsubsection*{Analysis of our semi-supervised method}
We evaluated the contribution of the semi-supervised learning in SegMatch by training only its fully-supervised branch, essentially turning it into OR-UNet. As discussed previously and shown in Table~\ref{stateart}, 
disabling the unlabelled pathway leads to a drop in terms of Dice and NSD scores thereby confirming the benefits of semi-supervised learning.

We also studied the effectiveness of varying the confidence threshold when generating the pseudo-labels as shown in Figure~\ref{cdplp}.
We observe that when the confidence threshold approaches 1.0, the model returns the worst segmentation performance.
When the threshold value is changed within the range of 
$[0.7,0.9]$,
the confidence threshold does not affect the model's performance significantly. However, it should be noted that further reducing the threshold value leads to a rapid decrease in the mean Dice score, which indicates that the quality of contributing pixels to the unsupervised loss may be more important than the quantity. This evaluation was conducted on the binary segmentation task using the \robustmis dataset, but a similar trend was observed in the multi-class segmentation task as well.

\subsubsection*{Augmentation strategy}
We also operated an ablation study on the weak and strong augmentation in SegMatch as tabulated in Table~\ref{abstdy}. 
First, we removed the adversarial augmentation, thereby only keeping handcrafted strong augmentations. Notably, we observed consistent results across different labelled data ratios. Specifically, when utilizing 17K additional unlabelled data, we found that the Dice score decreased by 3.1 pp compared to the full proposed SegMatch model.
This suggests that applying adversarial augmentations can prevent the learning saturation caused by hand-crafted strong augmentations on unlabelled data, thereby enabling the model to learn continuously.
When both the weak augmentation and adversarial augmentation are removed, the results drop by an additional 1.9 pp compared to only removing the adversarial augmentation, indicating that applying the weak augmentation function to the input image, which generates new and diverse examples for the training dataset, can enhance the segmentation performance.

To evaluate the effectiveness of the overall strong augmentation strategies, we replaced the strongly augmented images with the original images.
The resulting Dice score dropped by 5.1 pp compared to only removing the adversarial augmentation, and by 3.2 pp compared to removing both the weak and adversarial augmentation. The evidence that removing strong augmentation results in a greater decrease in performance than removing weak augmentation suggests that stronger perturbations are beneficial for learning with consistency regularization. 
Additionally, when strong augmentation is removed, the model is presented with the same input image from different views, which reduces the benefits of consistency regularization. In this scenario, pseudo-labelling becomes the primary technique for achieving better segmentation performance. Therefore, our results suggest that consistency regularization is more crucial than pseudo-labelling in our pipeline for improving segmentation performance.

We conducted an ablation study on augmentation strategies for multi-class segmentation using the CholecInstanceSeg dataset, as shown in Table \ref{abstdy-multi}. Even when trained on only 70\% of the data, our model outperformed the baseline OR-Unet \cite{isensee2020or}, which was trained on the full dataset as shown in Table \ref{multi-class-sta}, by 6.5 pp in Ch\_IoU and 6.84 pp in ISI\_IoU. This demonstrates the potential of our model in scenarios with limited annotated data. Consistent with results from the binary segmentation task, removing adversarial augmentation led to a noticeable drop in IoU scores, highlighting its role in preventing learning saturation from over-reliance on hand-crafted strong augmentations in unlabelled data. The results show that removing strong augmentation hurts performance more than removing weak augmentation, suggesting that, similar to binary segmentation, multi-class segmentation benefits from a greater input difference between branches for effective consistency regularization.

\subsubsection*{Adversarial augmentation analysis}
\begin{figure}
  \centering
  \begin{minipage}{0.4\columnwidth}
    \centering
    \includegraphics[width=\columnwidth]{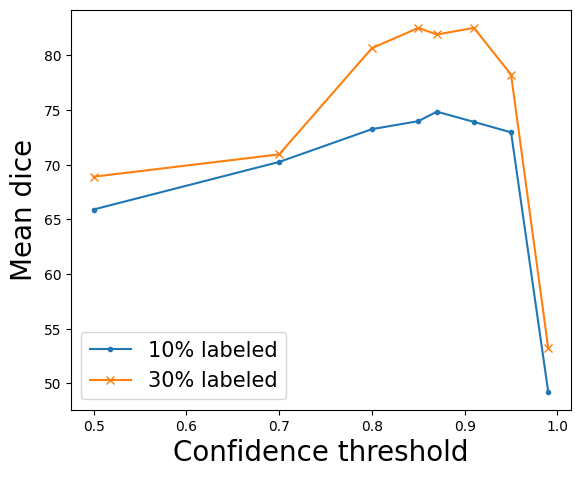}
    \caption{Mean Dice score produced by varying the confidence threshold for pseudo-labels}
    \label{cdplp}
  \end{minipage}\hfill
  \begin{minipage}{0.4\columnwidth}
    \centering
    \includegraphics[width=\columnwidth]{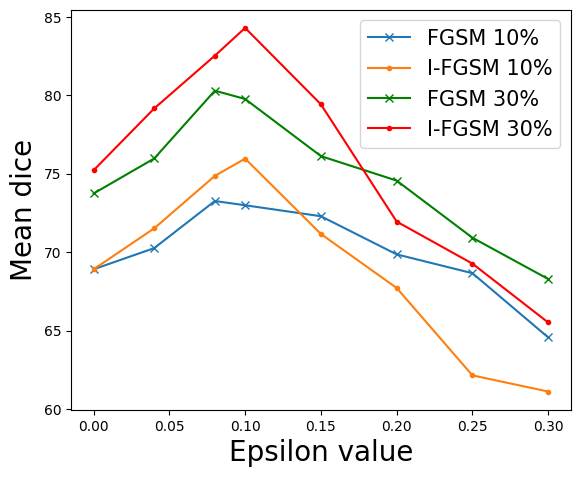}
    \caption{Optimal $\epsilon$ value enhances segmentation performance (as indicated by the peak in mean Dice score)}
    \label{equi}
  \end{minipage}
\end{figure}

\begin{figure*}[!t]
\centerline{\includegraphics[width=\linewidth]{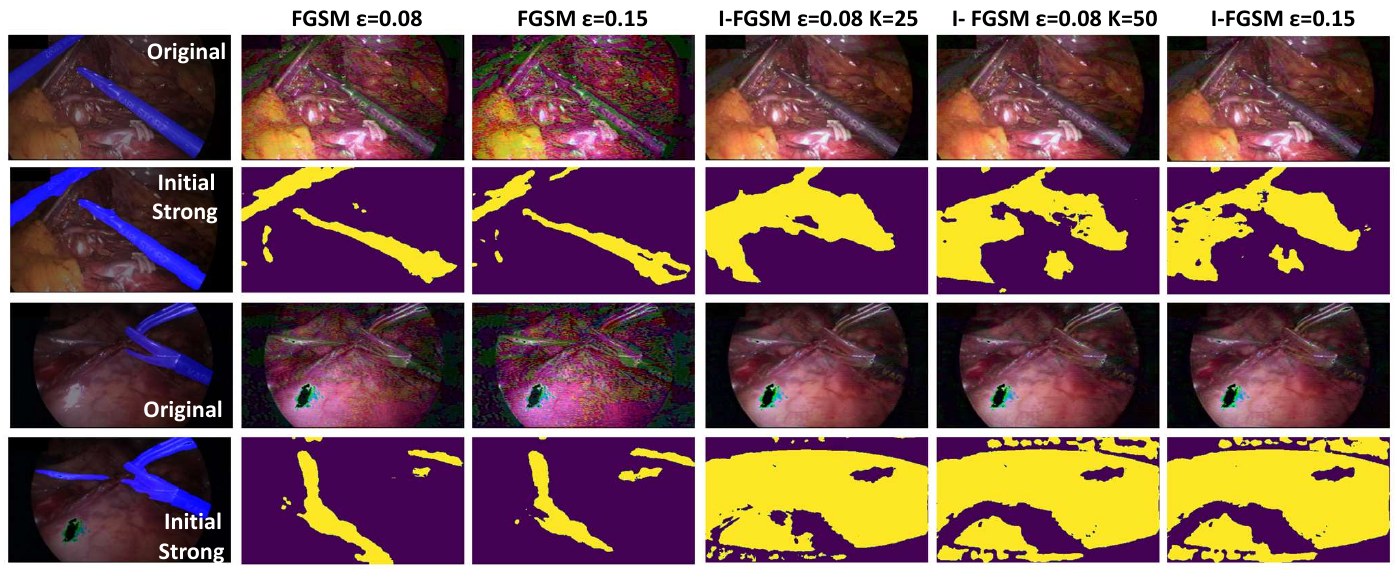}}
\caption{Examples showcase the impact of strong augmentation transform functions and adversarial augmentation on an original unlabelled image input to a model. The first column features the original image covered by its segmentation mask output from the model, as well as the strongly augmented image obtained via initial strong augmentation and its output segmentation mask. The 2-6 columns showcase adversarial images produced by I-FGSM with varying values of $\epsilon$ and $K$ (which becomes FGSM when $K=0$) to replace the original strongly augmented images for model parameter updating. The upper rows in the 2-6 columns display the adversarial images, while the bottom rows show the corresponding segmentation results produced by the model.}
\label{fgsmre}
\end{figure*}

We further evaluated the sensitivity of our results to changes in the maximum amplitude value $\epsilon$ of the adversarial perturbation as shown in Figure~\ref{equi} and qualitatively illustrated in Figure~\ref{fgsmre}.
When increasing from $\epsilon$ = 0.0 (i.e. no perturbation),
we observed a consistent pattern across different ratios of labelled data for both FGSM and I-FGSM.
Initially, as $\epsilon$ increased, the segmentation performance improved, reaching its peak at approximately $\epsilon$ = 0.08.
However, beyond this optimal point, the performance started to decline, indicating that stronger perturbations can enhance the model's performance only within a certain range. 

In this work, by precisely defining the acceptable range of perturbations, we restrict the perturbations within the $\epsilon$-neighbourhood, which ensures that the integrity of the instrument pixels is preserved while introducing subtle variations that aid in improved generalization of the model.
Comparing FGSM and I-FGSM, we found that I-FGSM showed superior performance to FGSM before reaching the optimal $\epsilon$ value.
However, after the optimal point, I-FGSM exhibited a more significant decrease in the model's performance compared to FGSM, which suggests that I-FGSM has a higher attack success rate than FGSM but becomes more harmful to the model when the attacking amplitude is large.

We also varied the number of I-FGSM iterations as shown in Table~\ref{tabeladvmethod} and qualitatively illustrated in Figure~\ref{fgsmre}.
Increasing the number of iterations in the I-FGSM attack shows a minor improvement consistent with expectations that it increases the attach success rate.
However, it is important to consider that this small improvement in segmentation performance through increased iterations comes with a trade-off in computational efficiency.

Additionally, we conducted experiments using various adversarial learning methods~\cite{carlini2017towards,rony2019decoupling} as shown in Table~\ref{tabeladvmethod}.
Our findings indicate that performance achieved by using one-step attack methods is consistently lower compared to our iterative strategy, which suggests when attempting to manipulate image samples to create adversarial examples, breaking the attack down into smaller steps can improve the overall success rate, as noted in previous research \cite{kurakin2018adversarial}.

The same ablation studies for adversarial attack methods were conducted on multi-class segmentation tasks using the CholecInstanceSeg dataset, as summarized in Table~\ref{tabeladvmethod-multi}. The results consistently align with those observed in binary segmentation, demonstrating that one-step attack approaches perform worse than our interactive adversarial augmentation. Notably, when the maximum perturbation amplitude $\epsilon$ is fixed, the number of iterations has a greater impact on the performance of multi-class segmentation compared to binary segmentation.

\begin{table}[htb!]
\centering
\caption{Segmentation performance of SegMatch on \robustmis when applying different adversarial attack methods, evaluating on binary segmentation task.}
\label{tabeladvmethod}
\resizebox{0.5\columnwidth}{!}{
\begin{tabular}{l c c c}
\toprule
\multirow{2}{*}{Adversarial method (within SegMatch)} & \multicolumn{2}{c}{Labelled: 1794(30\%)} & \\ 
\cmidrule{2-4}
  & Mean Dice & NSD \\ 
\midrule
C\&W \cite{carlini2017towards} & 72.9 \(\pm\) 2.22 & 70.4 \(\pm\) 2.13  \\

FGSM \cite{szegedy2013intriguing} & 75.8 \(\pm\) 2.16 & 79.6 \(\pm\) 2.05
 \\

DDN \cite{rony2019decoupling} & 78.9 \(\pm\) 2.21 & 77.2 \(\pm\) 2.52 \\\midrule

I-FGSM ($\epsilon=0.08, K=25$) & 84.3 \(\pm\) 2.3 & 80.2 \(\pm\) 1.9\\

I-FGSM ($\epsilon=0.08, K=50$) & 84.9 \(\pm\) 2.19 & 81.5 \(\pm\) 2.02\\
\bottomrule
\end{tabular}}
\end{table}

\begin{table}[htb!]
\centering
\caption{Segmentation performance of SegMatch on CholecInstanceSeg when applying different adversarial attack methods, evaluating on multi-class segmentation task.}
\label{tabeladvmethod-multi}
\resizebox{0.5\columnwidth}{!}{
\begin{tabular}{l c c c}
\toprule
\multirow{2}{*}{Adversarial method (within SegMatch)} & \multicolumn{2}{c}{Labelled: 12.57K (30\%)} & \\ 
\cmidrule{2-4}
  & Ch\_IoU & ISI\_IoU \\ 
\midrule
C\&W \cite{carlini2017towards} & 64.45 \(\pm\) 2.58 & 62.12 \(\pm\) 2.61  \\

FGSM \cite{szegedy2013intriguing} & 65.76 \(\pm\) 2.39 & 64.39 \(\pm\) 2.42 \\

DDN \cite{rony2019decoupling} & 67.24 \(\pm\) 2.68 & 66.56 \(\pm\) 2.75 \\\midrule

I-FGSM ($\epsilon=0.08, K=25$) & 72.12 \(\pm\) 2.49 & 69.04 \(\pm\) 2.52\\

I-FGSM ($\epsilon=0.08, K=50$) & 74.32 \(\pm\) 2.28 & 71.29 \(\pm\) 2.31\\
\bottomrule
\end{tabular}}
\end{table}

\begin{figure*}[htbp!]
\begin{center}
\includegraphics[width=\linewidth]{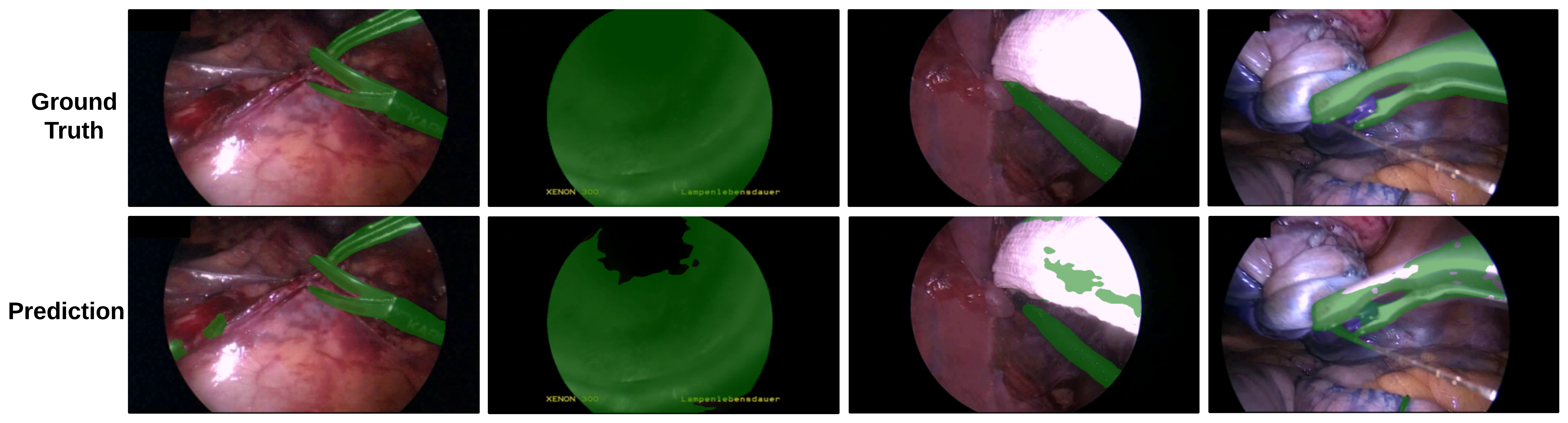}
\end{center}
\edtref{4}
\caption{\revnew{Failure cases of SegMatch's output. The first row shows the original image with the ground truth mask, while the second row overlays SegMatch predictions on the original image. Columns 1 to 4 illustrate the 1st, 2nd, 3rd, and 4th examples, respectively.}}
\label{figure:failcase}
\end{figure*}

\edtref{4}
\revnew{\subsection*{Failure cases analysis}}
\revref{1}{3}
\revnew{
We present failure cases in Figure \ref{figure:failcase}, highlighting challenges where SegMatch struggles to accurately segment surgical instruments. While adversarial and trainable augmentations allow flexible transformations to handle complex scenarios like overlapping tools or instruments obscured by tissue, the model faces difficulties with reflective surfaces (e.g., gauze in the 3rd example) and instruments resembling the tissue background (4th example), leading to false positives and imprecise pseudo-labels.
} 
\revref{3}{5}
\revnew{This implies noise in unlabelled data, such as poor image quality or artefacts, can reduce SegMatch's performance since consistency regularization may amplify errors. Additionally, out-of-distribution scenarios, such as the instrument-dominated image in the 2nd example, further challenge SegMatch, demonstrating its limitations in adapting to highly variable or ambiguous inputs despite leveraging unlabelled data effectively.}

%% file: 6_conclusion.tex
In this paper, we introduced SegMatch, a semi-supervised learning algorithm for surgical tool segmentation that achieves state-of-the-art
results across two of the most commonly used datasets in this field.
Our algorithm, SegMatch, was adapted from a simple semi-supervised classification algorithm FixMatch which combines consistency regularization and pseudo-labelling. 
During training, SegMatch makes use of a standard labelled image pathway and an unlabelled image pathway with training batches mixing labelled and unlabelled images.
The unlabelled image pathway is composed of two concurrent branches.
A weak augmentation branch is used to generate pseudo-labels against which the output of a strong augmentation branch is compared.
Considering the limitation of fixed handcrafted strongly augmentation techniques, we introduced adversarial augmentations to increase the performance of strongly augmented images.
We also highlighted the importance of considering equivariance and invariance properties in the augmentation functions used for segmentation.

Putting our work into its application context,
automatic visual understanding is critical for advanced surgical assistance but labelled data to train supporting algorithms are expensive and difficult to obtain.
\edtref{1}
\revnew{
SegMatch's ability to leverage unlabelled data makes it highly applicable to clinical environments, where appropriately annotated data is severely restricted due to time and budget constraints. Moreover, SegMatch's adaptability to varying levels of supervision aligns well with clinical workflows, where relying on partial annotations or weak labels is more feasible than the resource-intensive process of creating fully annotated datasets.

In robotic-assisted surgeries, our approach enables precise segmentation of surgical instruments during intraoperative procedures, facilitating real-time image analysis and delivering enhanced visual guidance to support surgical decision-making. Key applications include augmented reality for precise overlay placement, haptic feedback simulation to improve maneuver safety, and tissue mosaicking to broaden the surgical visual field. Furthermore, these advancements can contribute to surgical workflow optimization, objective skills assessment, camera calibration, and visual servoing, paving the way for improved accuracy, efficiency, and automation in complex surgical interventions.
}

\edtref{1}
\revnew{The principles underlying SegMatch are not confined to surgical instrument segmentation. Our trainable adversarial augmentation strategy could be equally beneficial in domains requiring precise segmentation of complex structures whereas annotations are limited and expensive to obtain.}
\revref{2}{1}
\revmod{We believe that simple but strong-performance semi-supervised segmentation learning algorithms, such as our proposed SegMatch, will not only accelerate the deployment of surgical instrument segmentation in the operating theatre but also be applicable to other domains lacking annotated data.}